\documentclass[letterpaper]{article}
\usepackage{ijcai15}
\usepackage{times}
\usepackage{savesym,multirow,url,xspace,graphicx,enumitem,color}
\usepackage[titletoc]{appendix}
\usepackage[tight]{subfigure}
\makeatletter
\newif\if@restonecol
\makeatother

\usepackage[lined, boxed, ruled, noend, commentsnumbered]{algorithm2e}
\usepackage{amsmath,amssymb,xfrac,amsthm}
\setitemize{noitemsep,topsep=1pt,parsep=1pt,partopsep=1pt, leftmargin=12pt}

\newtheorem{theorem}{Theorem}

\makeatletter

\newcommand{\Rmnum}[1]{\expandafter\@slowromancap\romannumeral #1@}
\makeatother
\usepackage{caption}

\usepackage{algpseudocode}
\usepackage{enumitem}
\usepackage{wasysym}

\DeclareMathOperator*{\argmax}{arg\,max}

\newcommand{\R}{\mathbb{R}}
\newcommand{\opt}{\textsc{VOpt}\xspace}

\newcommand{\elgreedy}{\textsc{NetExp}\xspace}
\newcommand{\actexplore}{\textsc{Explore}\xspace}
\newcommand{\actexploit}{\textsc{Exploit}\xspace}
\newcommand{\basedeg}{\textsc{Deg}\xspace}
\newcommand{\baseval}{\textsc{Val}\xspace}
\newcommand{\baserandom}{\textsc{Random}\xspace}

\newcommand{\citet}[1]{\citeauthor{#1} (\citeyear{#1})}

\newcommand{\groundSet}{V}
\newcommand{\selectSet}{S}
\newcommand{\featureSet}{X}
\newcommand{\task}{T}
\newcommand{\func}{f}
\newcommand{\graph}{G}
\newcommand{\edges}{E}
\newcommand{\quota}{Q}
\title{Information Gathering in Networks via Active Exploration}
\author{
Adish Singla\thanks{Adish Singla performed this research during an internship at Microsoft Research.}\\
ETH Zurich\\
adish.singla@inf.ethz.ch \\
\And
Eric Horvitz\\
Microsoft Research\\
horvitz@microsoft.com \\
\And
Pushmeet Kohli\\
Microsoft Research\\
pkohli@microsoft.com \\
\AND
Ryen White\\
Microsoft Research\\
ryen.white@microsoft.com \\
\And
Andreas Krause\\
ETH Zurich\\
krausea@ethz.ch \\
}
\begin{document}
\maketitle
\begin{abstract}
\vspace{-1mm}
How should we gather information in a network, where each node's visibility is limited to its local neighborhood? This problem arises in numerous real-world applications, such as surveying and task routing in social networks, team formation in collaborative networks and experimental design with dependency constraints. Often the informativeness of a set of nodes can be quantified via a submodular utility function. Existing approaches for submodular optimization, however, require that the set of all nodes that can be selected is known ahead of time, which is often unrealistic. In contrast, we propose a novel model where we start our exploration from an initial node, and new nodes become visible and available for selection only once one of their neighbors has been chosen. We then present a general algorithm \elgreedy for this problem, and provide theoretical bounds on its performance dependent on structural properties of the underlying network. We evaluate our methodology on various simulated problem instances as well as on data collected from social question answering system deployed within a large enterprise.
\end{abstract}
\vspace{-5mm}
\section{Introduction}\label{sec.introduction}
\vspace{-1mm}
Pioneering work of Stanley Milgram in the 1960's \cite{milgram1967small} provided evidence that individuals in a social network, possessing very limited knowledge of the whole network and only being able to access their close acquaintances, can effectively  route messages to distant target individuals in the network. Similar phenomena can be observed in many human-powered systems across a spectrum of examples, including the DARPA Red Balloon challenge\footnote{\url{http://archive.darpa.mil/networkchallenge/} \label{footnote:redballoon}} and co-authorship networks in academic communities. The challenges  of local knowledge and limited visibility often arise in various diverse computing systems. This includes querying in peer-to-peer networks constrained by the decentralized network design; information gathering on the internet by web crawlers or humans via following chains of linked documents \cite{white2011finding} constrained by the inability to directly discover new documents; and users seeking experts via friends in online social networks owing to the privacy constraints.  
{\em How can we build systems that {\em autonomously} explore networks under limited visibility for sake of information gathering?}

{\bf Information gathering.} We formalize these information gathering tasks as actively identifying a set of nodes in a network that maximize a set function quantifying their informativeness. Many natural objectives for this purpose satisfy {\em submodularity}, an intuitive diminishing returns condition (c.f., \citet{krause2011submodularity}). For instance, in the social Q\&A network (or web graph), the problem of finding experts in the network with the desired skills to answer the question (or the documents satisfying the information needs) can be cast as submodular function maximization \cite{el2009turning}. In the message routing problem in the Milgram's experiment, the utility function can be modeled as reduction in distance to the target, measured as the minimal distance from one of the selected nodes to the target. 

{\bf Local visibility of the network.} \looseness -1 Existing approaches for submodular function optimization are based on the key assumption that the ground set (of all nodes) is known in advance. With this assumption, greedy selection based on the marginal utilities of the nodes provides near-optimal solutions to the problem \cite{1978-_nemhauser_submodular-max}. However, having access to the entire network is unrealistic in many real-world applications for various reasons. For instance, due to privacy concerns, node visibility within social networks (such as Facebook or LinkedIn) is restricted to nodes we already connected to, and only these nodes can be target for routing tasks or posting a question. Even if the whole network is accessible (in a centralized system), the users may be more willing to respond and provide help to solve the task when routed through their acquaintances because of social incentives to help the peers and friends. The fundamental question is how to explore the local neighborhood of the network with the goal of maximizing the  utility function over the selected nodes.

\vspace{-2mm}
\subsection{Overview of our approach}\label{subsec.overview}
\vspace{-1mm}
We present a general approach to information gathering on networks under visibility constraints. In our model, the algorithm starts from an initial node (for instance, the individual in a social Q\&A network posting the question or seeking expertise for a task). New nodes become available for selection only once one of their neighbors has been already chosen. We model the local visibility constraints via limiting the number of hops within which the neighborhood of selected nodes becomes visible. Given this, the selection algorithm has to choose between exploring the network by selecting high degree nodes to expose new nodes and edges, or exploiting the currently accessible neighborhood by choosing nodes that provide maximal marginal utilities. Our main algorithmic contribution is a novel algorithm \elgreedy for this problem. We analyze its performance both in settings where no structural properties of the network are known, and in more specific settings capturing properties of real-world collaborative networks. Our main contributions are:
\begin{itemize}
\item A formal model of information gathering in networks with local visibility constraints, capturing several real-world applications, such as task routing in social networks.
\item A novel algorithm \elgreedy for this problem that actively explores the accessible local neighborhood to increase visibility of the unseen network, while at the same time exploiting the value of the information available in this neighborhood. We analyze the performance of our algorithm and provide theoretical bounds depending on the network structure.
\item Evaluation of our approach on data collected from a real-world application of task routing in a social Q\&A system deployed within a large enterprise, to show the practical applicability of our methodology.
\end{itemize}
\vspace{-2mm}
\section{Related Work}\label{sec.related}
\vspace{-1mm}
{\bf Search and navigation with local knowledge.}
Our work is inspired by the ideas of navigation and search in networks with local knowledge. The seminal work of \citeauthor{2000-stoc_kleinberg_small-world} (\citeyear{2000-stoc_kleinberg_small-world,2001-nips_kleinberg_small-world-dynamics}) addressed algorithmic questions of network formation and search strategies, to understand when short paths of acquaintances exists and are discoverable by local navigation.  In our work, we generalize the task of navigating to a target, to that of information gathering in a network. \citet{2001-_adamic_search-in-power-law-networks} and 
\citet{2003-_how-to-search-social-network} compared different local search strategies, for instance, following high-degree nodes
or using proximity in an organizational hierarchy in the context of an email network.  \citet{2014-jure-eric} tackled the challenge of navigating to a target in a geospatial network from a source using local knowledge of the network and proposed different navigation strategies -- based on degree, or being closer in distance to the target, as well as combining these two strategies.
Local search and team recruiting have also been studied in context of strategic participants via providing incentives \cite{2005-focs_query-incentive-network,2010-cikm_qin-social-networks}, which also formed the basis of recruiting participants in the Red Balloon challenge\textsuperscript{\ref{footnote:redballoon}}. 

{\bf Task routing.}
\looseness -1 \citet{2011-www_ryen_supporting_imx} developed ``IM-an-Expert", a synchronous  social Q\&A system, where a user can pose a question via instant messaging that is then routed to an ``expert" by the centralized system. Each posted question is characterized by the features and required skills; the users in the system have different expertise level, and the central system routes the question to the available experts. 
The main difference in our setting is the absence of any central authority, leading to the challenge of using local navigation to find the experts. More advanced variants of the social query model are proposed by \citet{2010-www_aardwark} and \citet{2008-_social-query-model}, however their proposed models  do not provide theoretical guarantees. \citet{2013-icwsm_crowd-powered-search} proposed ``SearchBuddies" that route questions to friends in online networks, but did not study the algorithmic aspects of this routing.  The challenge of locating experts in a decentralized manner has been studied \cite{2014-_besmira_task-routing,2010-icc_locating-experts}, however, no formal analysis has been provided for the proposed techniques.  \citet{2012-aamas_task-solving-and-routing} considered task routing for prediction tasks, by designing routing-based scoring rules and studying their truthfulness and efficiency in the equilibrium. However, we are not focusing on strategic aspects of the participants, and rather the algorithmic aspects of information gathering in networks.

{\bf Explore-exploit dilemma.}
 Many reinforcement learning  and online learning problems are associated with explore-exploit dilemma, and several solutions use the framework of multi-armed bandits (MAB) \cite{1985_lai_ucb_asymptotically}. Motivated by applications in social advertisement, \citet{2014-kdd_networked} considers social-MAB where the actions or arms are the users of the underlying social network. \citet{bnaya2013social} considers the application of social network search and targeted crawling by using MAB framework. One of the key differences in our work is that the ``exploration" refers to acquiring access to more nodes in the network that were previously inaccessible, and is very different from the usual notion of ``exploration" used in MAB literature.


\vspace{-2mm}
\section{Problem Statement}\label{sec.model}
\vspace{-1mm}
We now formalize the problem addressed in this paper.

{\bf Set of nodes and the network.}
\looseness -1 We consider a set of nodes (\emph{e.g.}, a population of people or users of a system) denoted by set $\groundSet=\{v_1, v_2,\dots,v_{|\groundSet|}\}$, of size $|\groundSet|$. There is an underlying network over the nodes, denoted by $\graph=(\groundSet,\edges)$. For now we will not make any assumptions about its structure. Instead we discuss this further during the analysis, as our bounds will depend on the specific assumptions that we assert.

{\bf Task and the utility function.}
We denote a task as $\task$,  associated with an initial node  $v_\task^o$. For instance, in a social Q\&A system, the task $\task$ could be a question, and  $v_\task^o$ is the user posting the question. We associate with each task $\task$ a function over the set of nodes, given by $\func_\task:2^\groundSet \rightarrow \R$, quantifying their informativeness/expertise.  Given a set of nodes $\selectSet$ selected for task $\task$, the utility achieved from this set is given by $\func_\task(\selectSet)$. We assume each $\func_\task$ to be \emph{non-negative},  \emph{monotone} (\emph{i.e.}, whenever $\selectSet \subseteq \selectSet' \subseteq \groundSet$, it holds that $\func(\selectSet) \leq f(\selectSet')$) and \emph{submodular}. Submodularity is an intuitive notion of diminishing returns, stating that, for any sets $\selectSet \subseteq \selectSet' \subseteq \groundSet$, and any given node $v \notin \selectSet'$, it holds that $f(\selectSet \cup \{v\}) - f(\selectSet) \geq f(\selectSet' \cup\{v\})-f(\selectSet')$. These conditions are general, and are satisfied by many realistic, as well as complex utility functions for information gathering \cite{krause2011submodularity,2012-survey_krause_submodular}.

{\bf Local visibility and connectivity of selected set.}
\looseness -1 We seek algorithms for the setting where the network is revealed incrementally as more nodes are selected. We denote the $l$-hop neighborhood of a node $v$ as $\mathcal{N}(v, l) \subseteq \groundSet$, to be the set of all nodes in the network that are connected to $v$ either directly or via at most $(l-1)$ intermediate nodes. For simplicity, we shall assume that $v$ is also included in $\mathcal{N}(v,l)$. For a set of nodes  $\selectSet$, we define its $l$-hop neighborhood as $\mathcal{N}(\selectSet,l) = \cup_{v \in \selectSet} \mathcal{N}(v,l)$. 
Since the network is unknown in the beginning, we seek an algorithm that incrementally selects nodes always within the $1$-hop neighborhood of already selected nodes. 

\begin{figure}[t]
\centering
\includegraphics[width=0.55\linewidth]{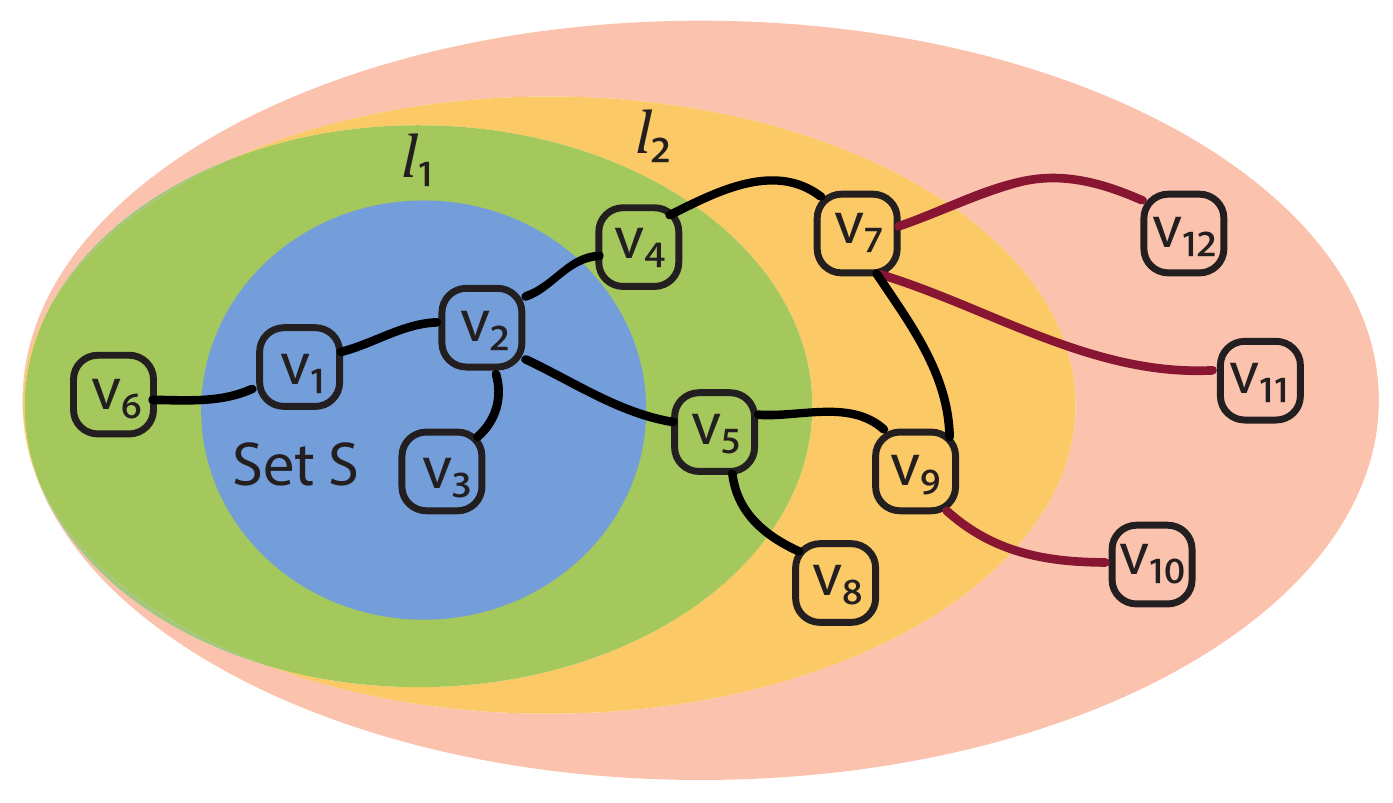}
\vspace{-1.2mm}
\caption{Illustration of local visibility constraints for $l_{deg}=2$, $l_{val}=2$. The currently selected set is given by $S=\{v_1, v_2, v_3\}$. The $1$-hop neighborhood of $S$, given by set $\{v_4, v_5, v_6\}$, is now accessible to the algorithm for selection. Given $l_{val}=2$, the algorithm can evaluate  $\func$ to compute marginal utilities for the set $\{v_4, v_5, v_6, v_7, v_8, v_9\}$. Given $l_{deg}=2$, the $1$-hop neighborhood of the nodes $\{v_4, v_5, v_6, v_7, v_8, v_9\}$ is visible to the algorithm in order to find the network they expose. For instance, the algorithm can compute that node $v_7$ exposes the set of nodes  $\{v_{9}, v_{11}, v_{12}\}$ }
\label{fig.fig-illustrate_visibility}
\end{figure}

In order to make informed decisions, we will assume that some information is being revealed to the algorithm while nodes are being selected. In particular, assuming we have already selected nodes $\selectSet$, for every node $v$ in  $N(\selectSet,l_{deg})$, its $1$-hop neighborhood $N(v,1)$ is visible. This essentially means that the algorithm has sufficient information in order to find the new network nodes that will be exposed if $v$ is added to the current set. We furthermore assume that the objective function $\func$ can be evaluated for every node $v$ in $N(\selectSet,l_{val})$. The constants $l_{deg}$ and $l_{val}$ will be specified later.  Figure~\ref{fig.fig-illustrate_visibility} illustrates this with a simple example for $l_{deg}=2$ and $l_{val}=2$.
In real-world social networks (such as Facebook or LinkedIn), the visibility is usually restricted to $l_{deg}=1$ and $l_{val}=1$ due to privacy settings.

{\bf Optimization problem.}
Consider a task $\task$ originating at any node $v^o$, and $\func$ its associated utility function. We seek to select a set of nodes  $\selectSet$ of  value at least $\quota$ in terms of utility. The process starts from the initial node $v^o$ and sequentially adds new nodes  maintaining the connectivity of  $\selectSet$. We are interested in designing an algorithm $\texttt{Alg}$ that has only $l_{deg}$ and $l_{val}$ visibility, and achieves $Q$ while minimizing the cost of the selected set. Formally, we define the optimization problem as follows:
\vspace{-2mm}
\begin{align*}\label{eq.opt}
\selectSet^* =& \operatorname*{arg\,min}_{\selectSet \subseteq \groundSet}|\selectSet| \tag{1} \\
\text{ s.t. } & \func(\selectSet) \geq \quota \tag{2} \\
& \selectSet \texttt{ is connected in } \graph \tag{3} \label{eq.opt.3} \\
& v^o \in \selectSet \tag{4} \label{eq.opt.4} \\
& \texttt{Algorithm Alg} \text{ has }  l_{deg}, l_{val} \text{ visibility } \tag{5} \label{eq.opt.5}
\vspace{-2mm}
\end{align*}
Note that due to constraint~\eqref{eq.opt.5}, an optimal solution to this problem is in fact not a fixed set, but a {\em policy} that specifies which next node to select as a function of the information revealed.
We also define a fixed optimal solution set $\opt$ that achieves value of $\quota$, not subject to any computational constraints nor the constraints given by \eqref{eq.opt.3}, \eqref{eq.opt.4} and \eqref{eq.opt.5}.

\vspace{-2mm}
\section{Our Algorithm \elgreedy}\label{sec.mechanism}
\vspace{-1mm}
We begin by describing the high level ideas behind our algorithm \elgreedy and then provide details. 

\vspace{-2mm}
\subsection{Overview of basic approach} \label{sec.mechanism.methodology}
\vspace{-1mm}
To present some of the key insights in designing  \elgreedy, let us first consider two extreme settings.

{\bf Special case: fully connected network.} Let us first consider the extreme setting where the network graph $\graph$ is fully connected. This, in turn, makes the constraints \eqref{eq.opt.3}, \eqref{eq.opt.4} and \eqref{eq.opt.5} for the optimization problem as posed in Section~\ref{sec.model} redundant. In this case, it reduces to an instance of standard submodular optimization \cite{1978-_nemhauser_submodular-max,krause2011submodularity}. Ignoring the computational constraints, the size of the fixed optimal solution set is given by $|\opt|$, as defined in Section~\ref{sec.model}. Obtaining this solution is intractable, however a greedy selection based on the marginal utilities of the nodes can provide a solution of size bounded by $O(|\opt| \cdot \ln(\frac{1}{\beta}))$ to achieve a utility of  $(1 - \beta) \cdot \quota$ where $\beta$ indicates a tolerance level. If $\func$ is integral, any $\beta>1/Q$ suffices to ensure complete coverage (\emph{i.e.}, exact value $Q$).

{\bf Building the network backbone.}  Now suppose the network is arbitrary. One approach would be to 
first build a connected ``backbone" of the network by selecting a set of nodes $\selectSet$ that covers the whole set $\groundSet$, \emph{i.e.}, $\mathcal{N}(\selectSet,1) =  \groundSet$. Once this ``backbone" is built, the whole network is visible and all the nodes are accessible to be selected by the algorithm. Subsequently, the problem again reduces to that of standard submodular optimization.
This ``backbone" is called a {\em connected dominating set} of the graph, and the size of minimal connected dominating set is given by $\gamma^c_G$. Computing such a minimal connected dominating set is intractable, but approximation algorithms are known \cite{1998-_khuller_cds-algorithmica}.
One naive approach to Problem~\ref{eq.opt} is hence to first build a complete ``backbone'', following by greedy selection of nodes based on marginal values. This approach guarantees a feasible solution, but it is obviously undesirable for large networks where  $\gamma^c_G$ might be very large. Furthermore, there may be some high value nodes closer to the initial node, and hence building the complete dominating set may be unnecessary.

\vspace{-2mm}
\subsection{Explore-exploit dilemma} \label{sec.mechanism.exploreexploit}
\vspace{-1mm}
The two basic extreme settings highlight two key regimes that any policy must be able to handle: {\em Exploit} an exposed (sub)-graph by selecting nodes to achieve value $Q$, and {\em explore} an unknown graph to expose nodes for selection. Our algorithm will interleave these steps.

{\bf Exploration of the network.} One natural choice for adding a node to the current set is to select one that provides maximal ``exposure" of the unconnected network.  The ``exposed" neighborhood of the selected set $\selectSet$ is the $l=1$ neighborhood excluding $\selectSet$, formally given by $\widetilde{\mathcal{N}}(\selectSet,1) = \mathcal{N}(\selectSet,1) \setminus S$.
Assuming we have already selected a set $\selectSet$, adding a new node  $v$ exposes an additional $\big|\widetilde{\mathcal{N}}(\selectSet \cup \{v\},1) \setminus \widetilde{\mathcal{N}}(\selectSet,1)\big|$ nodes that becomes available to the algorithm to be selected. This node $v$ may not provide  immediate value, however, the additional connections in the newly exposed network might be useful, potentially helping to discover and connect with the required high valued nodes. We call this greedy, value agnostic, choice ``exploration". 

\vspace{-3mm}
{\bf Exploitation of value.} The other natural choice of selecting the next node is to add the one that provides immediate value to the set in terms of maximal marginal utility value.
In context of an already selected set $\selectSet$, adding a new node $v$ provides a marginal gain in value, given by $\func(\selectSet \cup \{v\}) - \func(\selectSet)$. We call this choice ``exploitation", as it greedily maximizes utility given the currently exposed network. 

\begin{algorithm}[t!]
\nl {\bf Input}:\\
	\begin{itemize}
 		\item[$\Square$] Task: $\task$; Initial node: $v^o$; Utility function: $\func$\; 	
 		\item[$\Square$] Local visibility: $\{l_{deg}, l_{val}\}$\;
 		\item[$\Square$] Exploration rate: $\epsilon$; Value required: $\quota$; Tolerance: $\beta$\;
    \end{itemize}
\nl {\bf Output}: selected set $\selectSet$\;
\nl {\bf Initialize}:  $\selectSet=\{v^o\}$; $i=1$; $\epsilon^i=\epsilon$\;
\nl \While { $\func(\selectSet) < (1 - \beta) \cdot  \quota$ }{
\nl		    \If{$(\max_{v \in \widetilde{\mathcal{N}}(\selectSet,1)} |\mathcal{N}(\selectSet \cup \{v\},1)| = |\mathcal{N}(\selectSet,1)|)$}{
\nl				$\epsilon^i$ = 0; \label{alg.elgreedy.eps0} \Comment{(\emph{i.e.}, $\mathcal{N}(\selectSet, 1) = \groundSet$) Update $\epsilon^i$}\\
             }
\nl 		With prob. $\epsilon^i$, $a^i \leftarrow {\small \actexplore}$; else, $a^i \leftarrow {\small \actexploit}$; \label{alg.elgreedy.exp2} \\
\nl 		\eIf{$a^i = {\small \actexplore}$}{
\nl			$\Pi^* \!=\! \argmax_{\Pi^{\selectSet}_l: l \in [1 \ldots l_{deg}]} \frac{|\widetilde{\mathcal{N}}(\selectSet \cup \Pi^\selectSet_l,1) \setminus \widetilde{\mathcal{N}}(\selectSet,1)|}{l}$ \label{alg.elgreedy.chain1}\;
\nl			$\selectSet = \selectSet \cup \Pi^*$\;
\nl			\If{$l_{deg} = 1$}{
\nl				 {\small Randomly pick} $v^* \in \widetilde{\mathcal{N}}(\selectSet \cup \Pi^*,1) \setminus \widetilde{\mathcal{N}}(\selectSet,1)$; \label{alg.elgreedy.random}
\nl				$\selectSet = \selectSet \cup \{v^*\}$\;
			}
		}
		{
\nl			$\Pi^* = \argmax_{\Pi^{\selectSet}_l: l \in [1 \ldots l_{val}]} \frac{\func(\selectSet \cup \Pi^\selectSet_l) - \func(\selectSet)}{l}$ \label{alg.elgreedy.chain2}\;
\nl			$\selectSet = \selectSet \cup \Pi^*$\;
		}
\nl		$i=i+1; \epsilon^i = \epsilon^{i-1}$\;
	}
\nl {\bf Output}: $\selectSet$\\
\caption{Algorithm \elgreedy}
\label{alg.elgreedy} 
\end{algorithm}
%
\vspace{-2mm}
\subsection{Algorithm \elgreedy} \label{sec.mechanism.algorithm}
\vspace{-1mm}


{\bf Interleaving network exploration with exploitation.} The key idea is to interleave the two choices of network exploration and exploitation. While the exploration step will continue towards building the connected dominating set (that we need in the worst case anyways as per results in Theorem~\ref{theorem.general.lower}), the exploitation step will greedily select the nodes to maximize the information gathered and terminate the algorithm as soon as the desired level of utility is achieved. Our algorithm \elgreedy is illustrated in Algorithm~\ref{alg.elgreedy}. It turns out that this simple approach allows us to derive tight theoretical bounds on the performance, and it also runs quite efficiently for various problem instances.  \elgreedy  interleaves exploration and exploitation with $\epsilon$ probability, as illustrated in Step~\ref{alg.elgreedy.exp2} of Algorithm~\ref{alg.elgreedy}. The $\epsilon$ is constant for the procedure and is provided as input. If further prior information about the network properties or the distribution of features is available, this parameter can be tuned (for instance, whether to do more exploration or more exploitation based on such properties). In Algorithm~\ref{alg.elgreedy}, we simply use a constant $\epsilon$,  and when the full network is exposed (or dominating set for the network is already built), we set $\epsilon=0$ (see Step~\ref{alg.elgreedy.eps0}).

{\bf Look-ahead during exploration}
As noted above, the exploration steps of the algorithm aim towards building the connected dominating set. Negative results from \citet{1998-_khuller_cds-algorithmica} show that to effectively build connected dominating sets, adding one node at a time based on the criterion of maximal ``exposure" is not sufficient, as such a greedy approach may need upto $\Omega(|\groundSet|)$ nodes to build a connected dominating set. Our idea is based on the intuition used by \citet{1998-_khuller_cds-algorithmica} on how to effectively add upto two nodes, when $l_{deg} \geq 2$,  in a way to be able to build efficient connected dominating sets.  We generalize the idea of one-step look ahead for $l_{deg} = 2$ to that of doing a $l-1$ step look ahead for $l_{deg} = l$ during exploration. To formalize this, we introduce the concept of a {\em chain of length $l$}. Consider the current selected set $\selectSet$. We define an $l$-chain from $\selectSet$, denoted by $\Pi^\selectSet_l$ as an ordered set satisfying the following constraints:
\vspace{-2mm}
\begin{align*}
&\Pi^\selectSet_1 \in \widetilde{\mathcal{N}}(\selectSet,1);\\
&\Pi^\selectSet_2 \in \widetilde{\mathcal{N}}(\selectSet \cup \Pi^\selectSet_1, 1) \setminus \widetilde{\mathcal{N}}(\selectSet,1);\\
&\Pi^\selectSet_{i_{\geq3}} \in \widetilde{\mathcal{N}}(\selectSet \cup \Pi^\selectSet_{i-1},1) \setminus \widetilde{\mathcal{N}}(\selectSet \cup \Pi^\selectSet_{i-2},1);
\vspace{-3.5mm}
\end{align*}

In Step~\ref{alg.elgreedy.chain1}, all possible chains of length $1$ to $l$ are enumerated to find the set of nodes to add at every iteration. While only considering chains is not required for the analysis, it restricts the search space. Nevertheless, the number of chains to consider is exponential in the lookahead. However, this is not an issue for practical settings where  $l \leq 2$ usually.

{\bf Look-ahead during exploitation}
The similar idea of look-ahead as used above is also employed during the exploitation step, and is done in Step~\ref{alg.elgreedy.chain2}. This look-ahead can also be thought of as moving in the direction of the utility function's (discrete) ``gradient".

{\bf The case of $l_{deg} = 1$.}
Recently, \citet{2012_power-of-local-info} introduced a simple randomization technique to efficiently build up connected dominating sets when ``look-ahead" is not possible, \emph{i.e}, for the case $l_{deg} = 1$. Specifically, after a node $v$ is added, another randomly selected node from its neighborhood $\mathcal{N}(v,1)$ that is newly exposed is also added. We use the same trick in \elgreedy, in Step~\ref{alg.elgreedy.random} of Algorithm~\ref{alg.elgreedy}.

\vspace{-3mm}
\section{Performance Analysis}\label{sec.analysis}
\vspace{-2mm}
We now analyze the performance of \elgreedy. The proofs and details are available in the extended version of the paper  \cite{2015-arxiv_singla_netexp_longer}.
 

\vspace{-2mm}
\subsection{Analysis for general settings}\label{sec.analysis.general}
\vspace{-1mm}
We first analyze \elgreedy for general settings, without any assumptions on the network structure or the utility function. We will state our results in terms of two important network properties: \emph{i)}  the maximum degree of any node in the graph $\graph$, which we denote as $\Delta_\graph$, and \emph{ii)} smallest size of any \emph{connected} dominating set $\gamma_\graph^c$ as defined in Section~\ref{sec.mechanism.methodology}.  
We start with analyzing the setting where $(l_{deg}=2$, $l_{val}=1)$ (similar results hold for $l_{deg} \geq 2$, $l_{val} \geq 1$ as well):
\vspace{-1mm}
\begin{theorem}\label{theorem.general.l2}
For $l_{deg}=2$, $l_{val}=1$, \elgreedy terminates with set $\selectSet$ achieving  utility of at least $(1 - \beta) \cdot \quota$, satisfying the constraints of \eqref{eq.opt.3}, \eqref{eq.opt.4} and \eqref{eq.opt.5} for Problem~\ref{eq.opt}, with the following bound on the size of $\selectSet$ in expectation (over the coin flips), given by $\mathbb{E}[|\selectSet|] \leq$
\vspace{-2mm}
\begin{align*}
\Big(\frac{1}{\epsilon} \!\cdot\! \big(2 + 2\ln(\Delta_\graph) \big) \!\cdot\!  \gamma^c_\graph \Big) + \Big(\frac{1}{1- \epsilon} \cdot |\opt| \cdot \ln(\frac{1}{\beta})\Big)
\end{align*}
\vspace{-4mm}
\end{theorem}

Common social networks only satisfy $l_{deg}=1$ (\emph{i.e.}, nodes can see their friends' friends). For this more challenging case we can still prove the following (slightly weaker) result:
\begin{theorem}\label{theorem.general.l1}
For $l_{deg}=1$, $l_{val}=1$,\elgreedy terminates with set $\selectSet$ achieving utility of at least $(1 - \beta) \cdot \quota$, satisfying the constraints of \eqref{eq.opt.3}, \eqref{eq.opt.4} and \eqref{eq.opt.5} for Problem~\ref{eq.opt}, with the following bound on the size of $\selectSet$ in expectation (over the coin flips) that holds with probability at least $1 - e^{-\gamma^c_G}$, given by
\vspace{-2mm}
\begin{align*}
\mathbb{E}[|\selectSet|] \!\leq\! \Big(\frac{1}{\epsilon} \!\cdot\! \big(4 + 2\cdot\ln(\Delta_\graph) \big) \!\cdot\!  \gamma^c_\graph \Big)\! +\! \Big(\frac{1}{1- \epsilon} \!\cdot\! |\opt| \!\cdot\! \ln(\frac{1}{\beta})\Big)
\end{align*}
\vspace{-4mm}
\end{theorem}

The proof of Theorems~\ref{theorem.general.l2} and~\ref{theorem.general.l1} involves separately analyzing the sequence of explore and exploit actions during the execution of the procedure, and ensuring that interleaving these sequences preserves the analysis. 

For the general setting, we also have the following lower bound, showing that the dependency on $\gamma^c_G$ is unavoidable.
%
\vspace{-1mm}
\begin{theorem}\label{theorem.general.lower}
For any bounded $l_{deg}$ and $l_{val}$, there exists a problem instance for which any feasible policy will need to select a set of size at least  $|\selectSet| \geq \max{} \Big(\gamma^c_\graph,   \text{ }   |\opt|\Big)$
\vspace{-1mm}
\end{theorem}
The proof of Theorem~\ref{theorem.general.lower} follows from the arguments of Section~\ref{sec.mechanism.methodology} and by crafting a worst-case instance of graph structure and distribution of node values for any given policy.

\vspace{-2mm}
\subsection{Analysis for realistic settings}\label{sec.analysis.pa}
\vspace{-1mm}
We now analyze the results for the specific setting motivated by real-world collaborative networks  such as co-authorship networks in academic communities.

{\bf Feature distribution and network structure.}
The nodes are associated with subsets of a feature set $\featureSet$ of size $|\featureSet|$. Considering the specific domain of co-authorship networks, a feature $x \in \featureSet$ could, \emph{e.g.}, be an indicator variable denoting whether a user has published a paper in an AI conference. For each feature $x \in \mathcal{X}$, let the set of nodes possessing that feature be given by  $\groundSet_x$.  We consider each feature $x \in \mathcal{X}$ as a ``social dimension" that induces a network given by $\graph_x=(\groundSet_x,\edges_x)$. In this model we assume that each network $\graph_x$ is formed by a preferential attachment process \cite{1999-science_emergence-scaling} leading to a power law distribution of the degrees. The final network that we observe is obtained by an overlay of these networks, given by $\graph = \cup_{x \in \mathcal{X}} \graph_x$. For a node $v$, we denote the value of its feature $x \in \featureSet$ as $x_v$ and  is proportional to the rank order dictated by the degree of node $v$ in graph $\graph_x$. This notion of feature values essentially captures the ``authority" of a node over a particular feature (for instance, a node gains expertise in ``AI" if it has high degree of connections for this particular network).

{\bf Characteristics properties of utility function.}
For a given task $\task$, we consider a separable utility function given by: $\func(\selectSet) = \sum_{x \in \featureSet} w^x \cdot \func^x(\selectSet)$,  where $w^x$ denotes the weight of function $f^x$. Here, the function $\func^x$ depends only on nodes' features  $x$.  
The maximum value of function $\func^x(\selectSet) = 1$ is achieved after including a top valued node with feature $x$. 

Theorem~\ref{theorem.pa} states the main result for these settings, and is proved using the results on the navigation properties of ``small-world" networks \cite{2003-_mathematical-results-pa,2012_power-of-local-info}. Compared to the general settings of Theorem~\ref{theorem.general.l2} and~\ref{theorem.general.l1}, the ``small-world" networks are much easier to navigate as can be seen in the polylogarithmic bounds.
\vspace{-1mm}
\begin{theorem}\label{theorem.pa}
Consider $l_{deg}=1$, $l_{val}=1$ and a task $T$ originating from user $v^o$ possessing a feature $x^o$. With probability of at least $1 - o(1)$, \elgreedy terminates with set $\selectSet$ achieving  utility of at least $(1 - \beta) \cdot \quota$, with the following bound on the size of $\selectSet$ in expectation, given by  $\mathbb{E}[|\selectSet|] \leq$
\vspace{-2mm}
\begin{align*}
\Big(\frac{1}{\epsilon} \!\cdot\! O\big(\ln^4(|\groundSet_{x^o}|)\big)\Big)\! +\! \Big(\frac{1}{1 - \epsilon} \!\cdot\! O\big(\sum_{x \in \featureSet} \ln^4(|\groundSet_x|) + |\opt| \big)\Big) 
\vspace{-7mm}
\end{align*}
\end{theorem}
\vspace{-2mm}

Next, we state the lower bound in Theorem~\ref{theorem.pa.lower}, which follows from the expected diameter of the graph obtained from preferential attachment \cite{2004-_scale-free-preferential}.
\vspace{-1mm}
\begin{theorem}\label{theorem.pa.lower}
For any bounded $l_{deg}$ and $l_{val}$, there exists  a problem instance for which any feasible policy will need to select a set of size at least $\mathbb{E}[|\selectSet|] \geq  O\Big(\frac{\ln(|\groundSet|)}{\ln\ln(|\groundSet|)}\Big)$
\vspace{-3mm}
\end{theorem}

\section{Experimental Evaluation}\label{sec.experiments}
\vspace{-1.5mm}
We now report on the results of our experiments.
\vspace{-1mm}

\begin{figure*}[t!]
\centering
   \subfigure[Erd\H{o}s-R\'enyi model]{
     \includegraphics[width=0.26\textwidth]{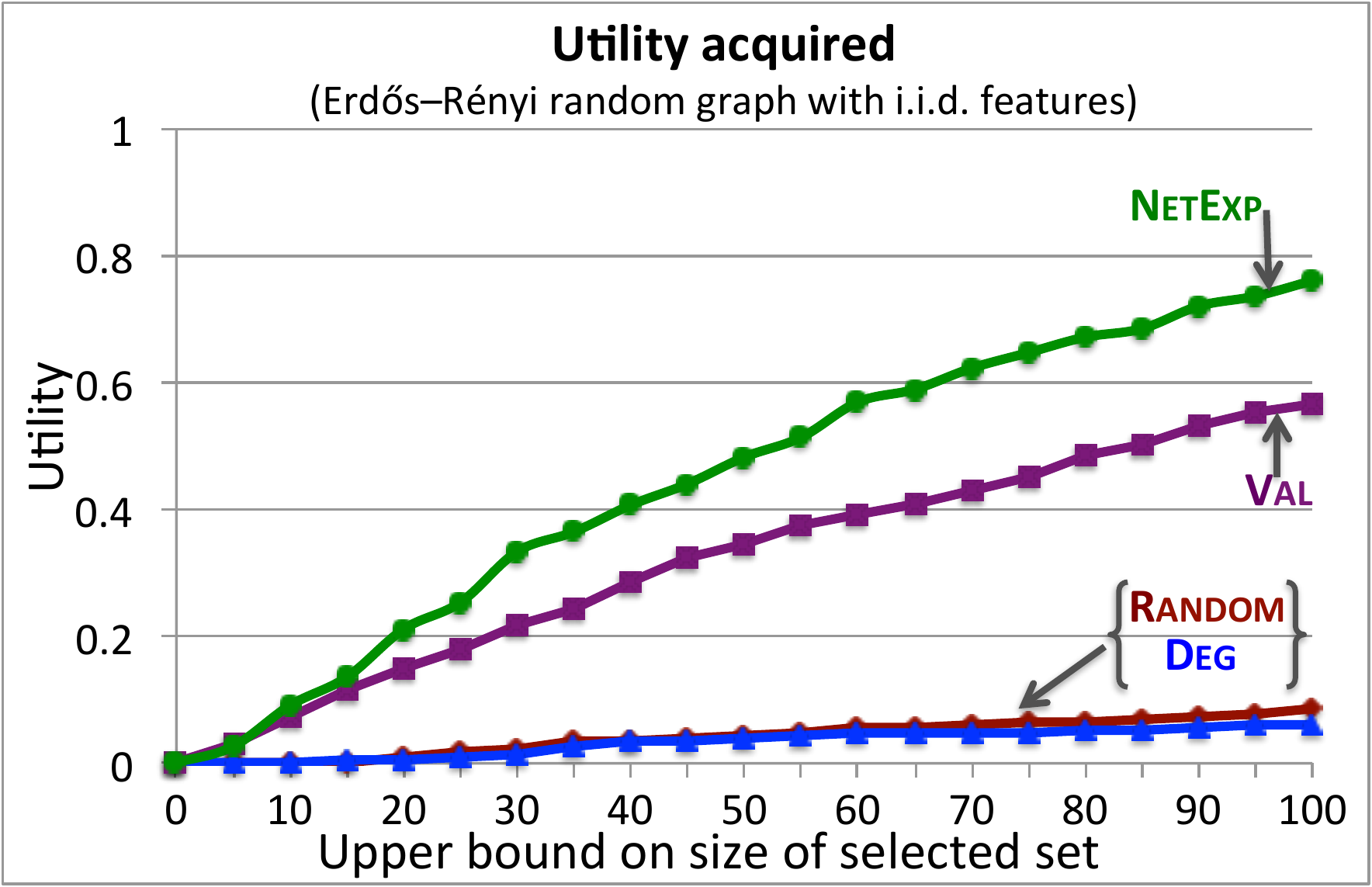}
     \label{fig.result_syn_random_vary-cost_util}
   }
   \subfigure[Preferential Attachment model]{
     \includegraphics[width=0.26\textwidth]{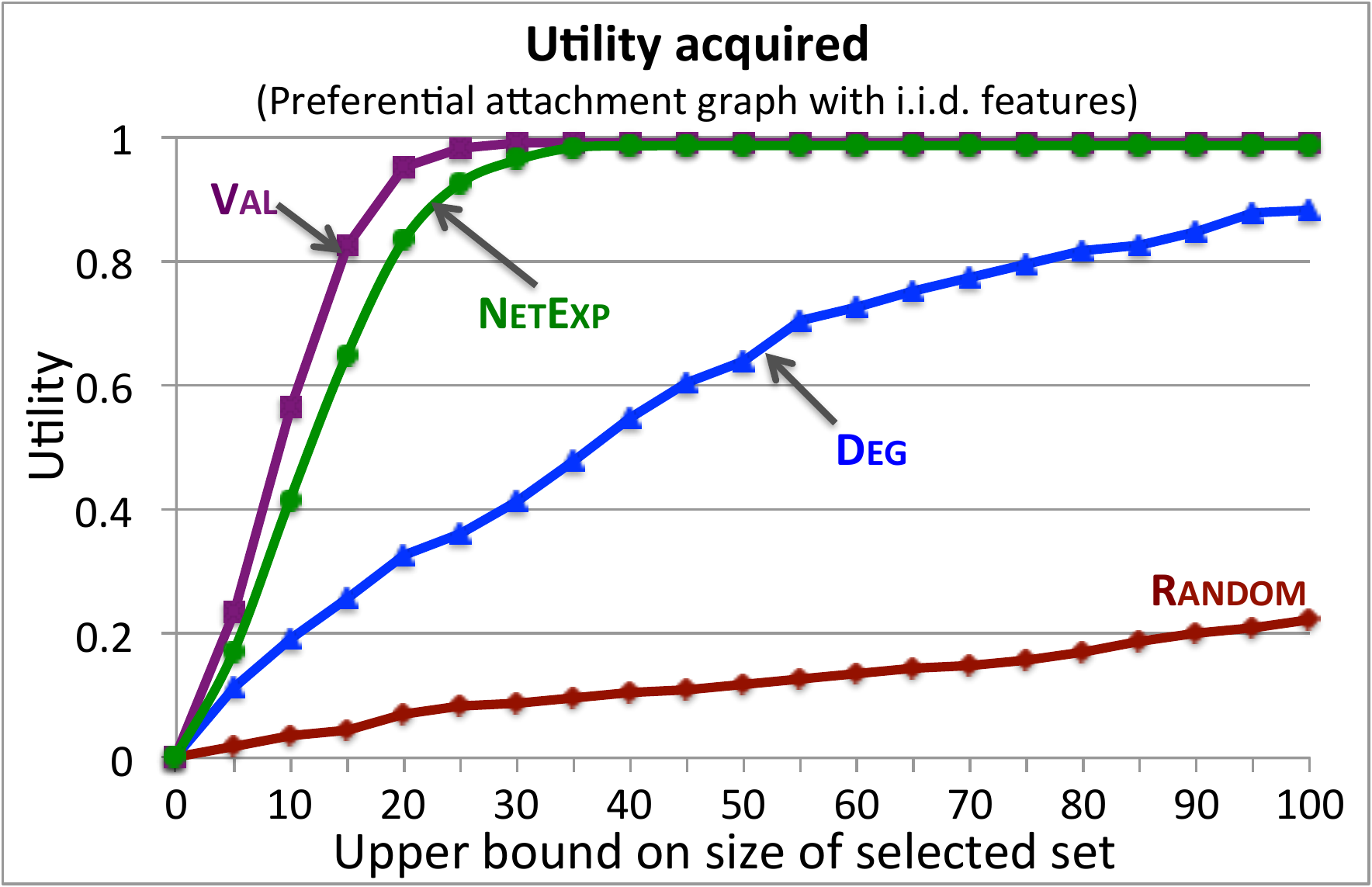}
     \label{fig.result_syn_pa_vary-cost_util}
   }
   \subfigure[Real-world social Q\&A system]{
     \includegraphics[width=0.26\textwidth]{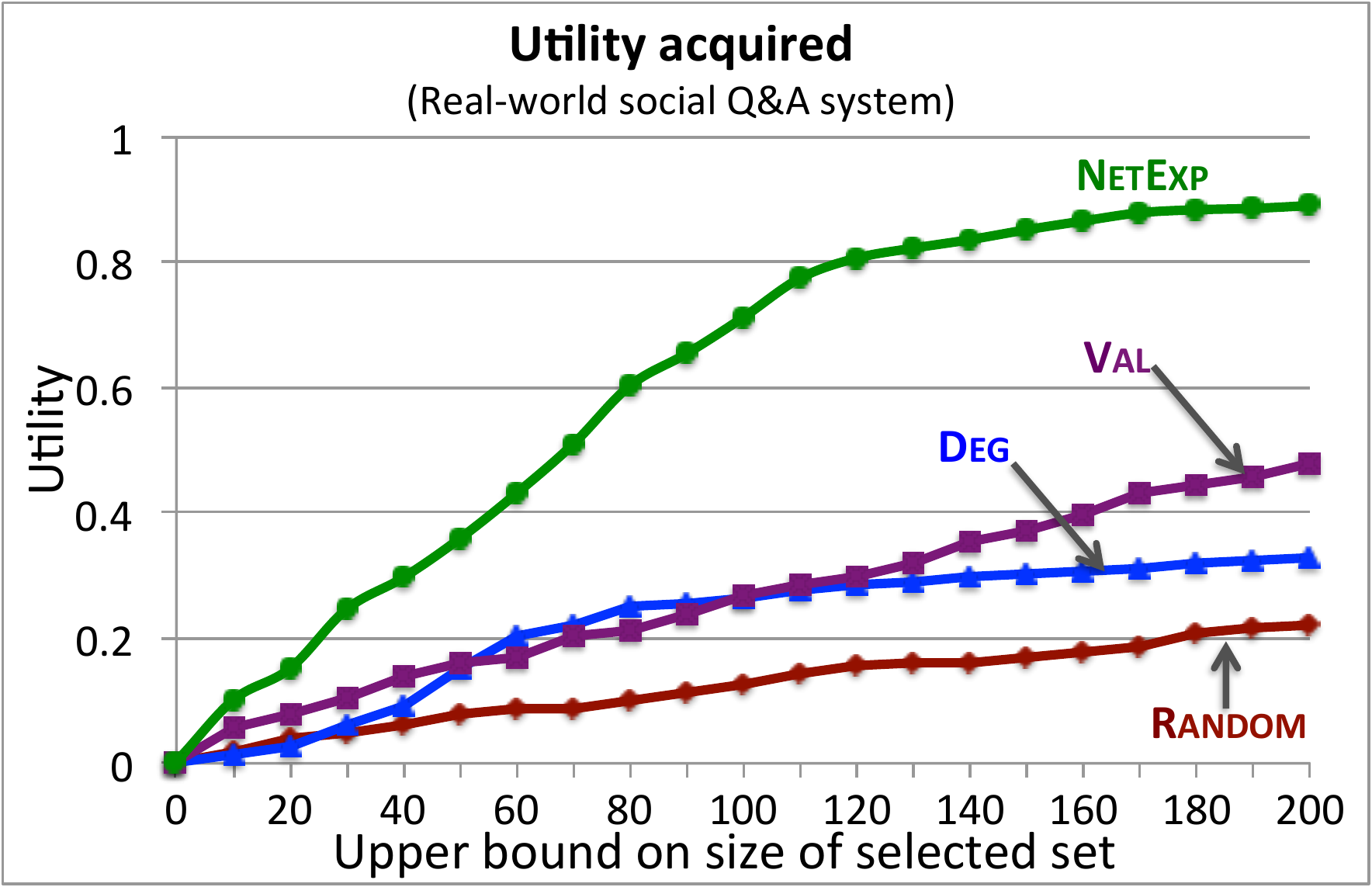}
     \label{fig.result_real_imx_vary-cost_util}
   }
   \subfigure[Exposed network]{
     \includegraphics[width=0.26\textwidth]{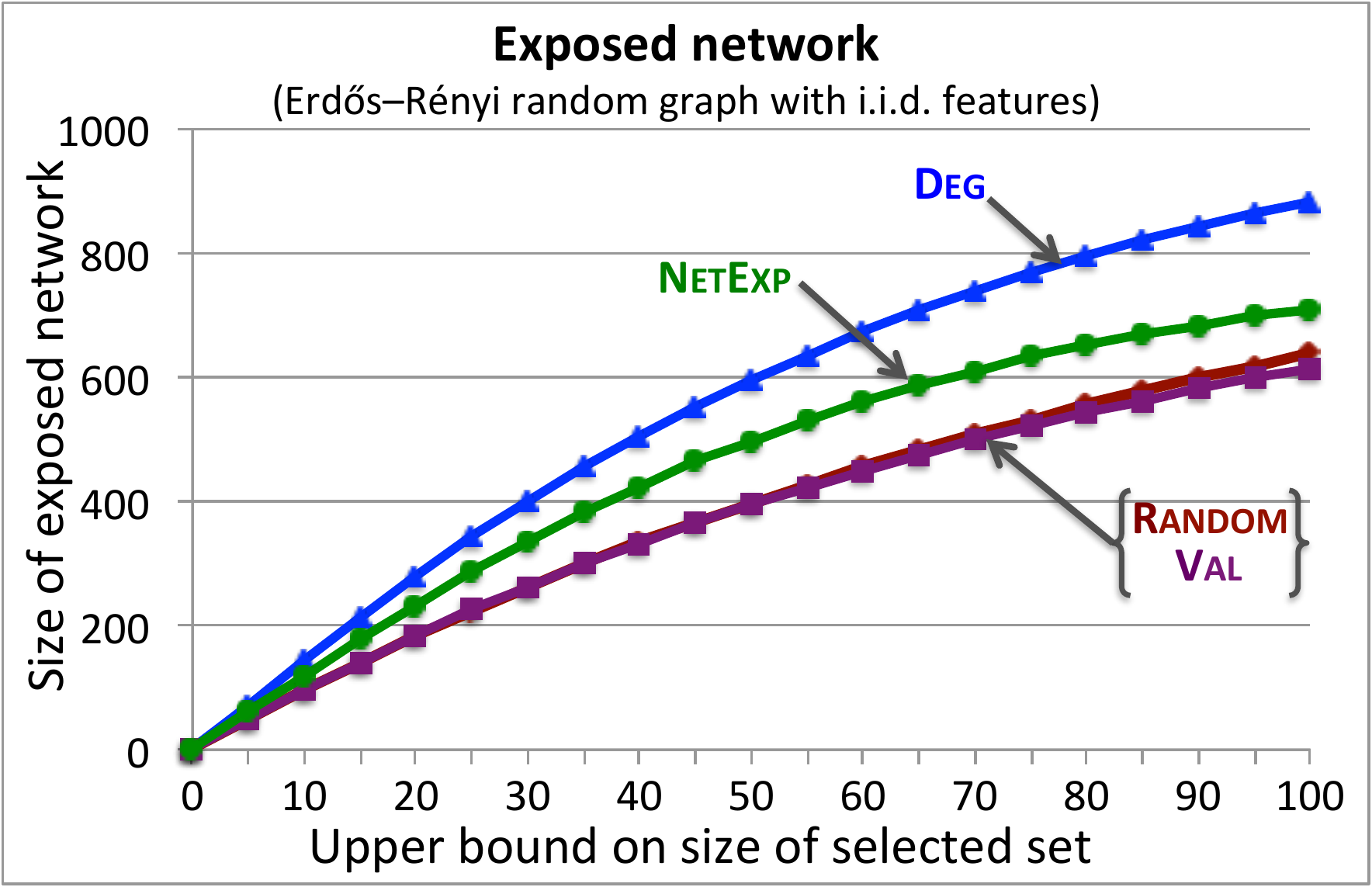}
     \label{fig.result_syn_random_vary-cost_visibility}
   }
   \subfigure[Varying fraction of  experts]{
    \includegraphics[width=0.26\textwidth]{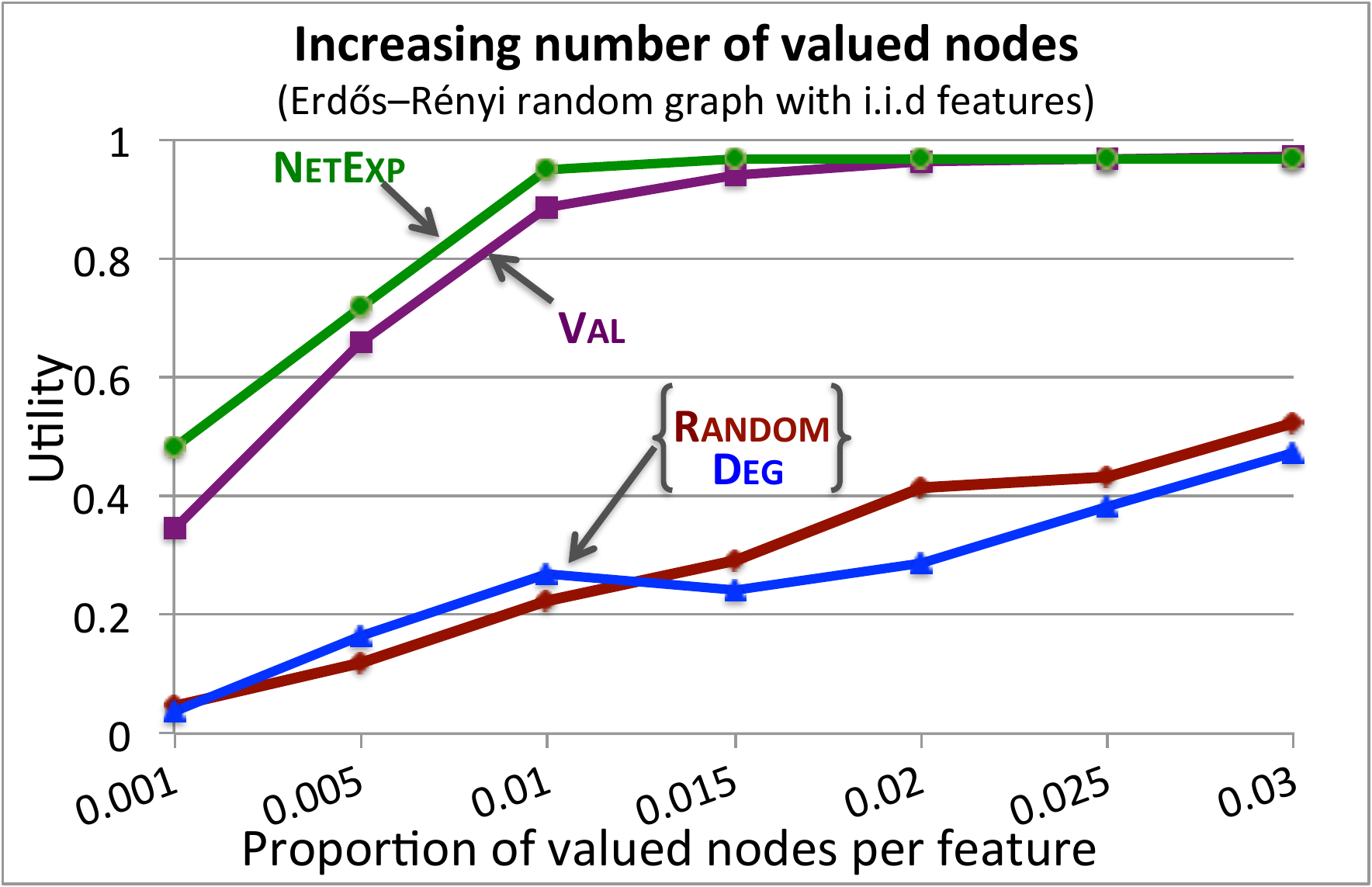}
     \label{fig.result_syn_random_vary-probExpert}
   }
   \subfigure[Varying $\epsilon$ for \elgreedy]{
     \includegraphics[width=0.26\textwidth]{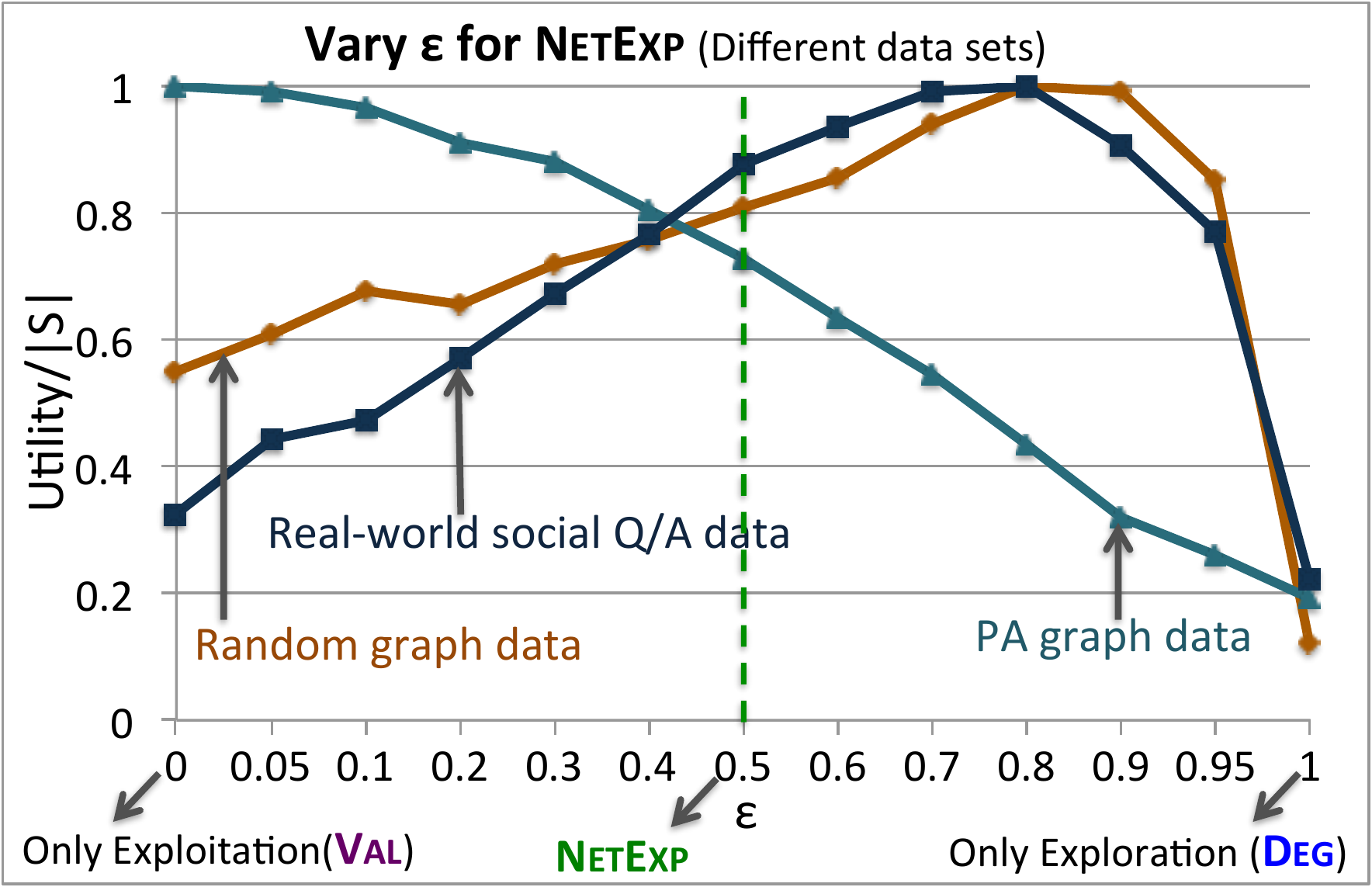}
     \label{fig.result_vary-epsilon}
   }
\vspace{-2.5mm}
\caption{Figure~\ref{fig.result_syn_random_vary-cost_util}, \ref{fig.result_syn_pa_vary-cost_util} and~\ref{fig.result_real_imx_vary-cost_util} shows the utility acquired as the size of the selected set $\selectSet$ increases. Results show robustness of \elgreedy for different network structures and feature distributions.  For the Erd\H{o}s-R\'enyi graph model, Figure~\ref{fig.result_syn_random_vary-cost_visibility} shows the exposed network for different algorithms and Figure~\ref{fig.result_syn_random_vary-probExpert} shows the results of varying the proportion of valued nodes. Figure~\ref{fig.result_vary-epsilon} shows the results of varying the $\epsilon$ for \elgreedy, showing the robustness of the \elgreedy compared to explore-only ($\epsilon=1$) and exploit-only ($\epsilon=0$) methods.}
\label{fig.resuts1}
\vspace{-1mm}
\end{figure*}

\vspace{-1mm}
\subsection{Experimental Setup}
\vspace{-1mm}
{\bf Benchmarks and Metrics.} We compare the performance of our procedure \elgreedy against the following three baselines: \emph{i}) \baserandom is a trivial baseline that randomly selects users from the exposed network to add to the set; ; \emph{ii}) \basedeg is equivalent to running  \elgreedy with $\epsilon = 1$; and \emph{iii}) \baseval is equivalent to running \elgreedy with $\epsilon = 0$. The primary metric that we compare is the utility achieved as a function of the size of the selected set $\selectSet$. We further carried out experiments by varying $\epsilon$, varying the distribution of node values and comparing the size of the exposed network.

{\bf Utility function and features.}
We consider a feature set $\featureSet$ of size $|\featureSet|$ associated with the nodes $\groundSet$, and denote the value of feature $x \in \featureSet$ for node $v \in \groundSet$ as $x_v$. For a given task $\task$, we consider separable utility functions per feature and can write $\func(\selectSet) = \frac{1}{\sum_{x \in X} w^x} \sum_{x \in \featureSet} w^x \cdot \func^x(\selectSet)$, with weights given by $w^x \in \{0,1\}$. We consider a probabilistic notion of utility derived from the feature values and define a submodular utility function  per feature $\func^x(\selectSet) = \Big(1 - \prod_{s \in \selectSet} (1 - x_s)\Big)$, inspired by \citet{el2009turning}.

{\bf Implementation choices.} We considered a realistic setting of $l_{deg}=1$ and $l_{val}=1$ visibility. For running different algorithms, we considered value quota $Q = 1$, with tolerance of $\beta=0.05$.  We used $\epsilon=0.5$ for \elgreedy across all of the datasets without any further tuning. For the implementation of the \elgreedy, we report results where we skipped the Step~\ref{alg.elgreedy.random} of Algorithm~\ref{alg.elgreedy} of adding a random user after every exploration step for $l_{deg}=1$ case. While this step is useful to handle the worst case scenarios of the graph configuration to achieve desirable bounds, for practical purposes, it did not provide any additional benefit or performance gain. 

\vspace{-2mm}
\subsection{Datasets}
\vspace{-1mm}
We performed experiments with three different datasets. 

{\bf Erd\H{o}s-R\'enyi random graph.}
We created random graphs using the random graph model of \citet{erd1959random}, with ground set of size $|\groundSet| = 1000$ and the probability of an edge existing between any two nodes given by $p_{edge} = 0.01$, a threshold probability above which the graph is connected with high probability. We then considered a total of $5$ features (\emph{i.e.}, $|\featureSet| = 5)$, distributed \emph{i.i.d.} across the nodes with probability given by $p_{val} = 0.001$. Note, that this probability is essentially equivalent to having one non-zero valued node in the set  per feature on expectation. For these \emph{non-zero} valued nodes per feature, the actual values of the features are then uniformly sampled and scaled to lie in the range $(0, 1]$.  We also vary  $p_{val}$ used by results in  Figure~\ref{fig.result_syn_random_vary-probExpert}.  We generated a set of the tasks, each associated with $3$ required features out of $5$ (\emph{i.e.}. $3$ functions have non-zero weights $w^x$). Distribution of tasks is generated by a random sampling of the required features, as well as the initial user $v^o$ per task.

\vspace{-0.5mm}
{\bf Preferential attachment graph per skill.}
As a second dataset, we used settings similar to those discussed in Section~\ref{sec.analysis.pa}. We considered a ground set of size $|\groundSet| = 100,000$ and total of $10$ distinct features (\emph{i.e.}, $|\featureSet| = 10)$. The features are distributed \emph{i.i.d.} with probability $0.2$ (\emph{i.e.}, on the expectation, every node is associated with two features).  The distribution of the tasks is generated similar to the first dataset, with three randomly selected relevant skills per task.

\vspace{-0.5mm}
{\bf Real-world social Q\&A system.}
As a third study, we used a real-world dataset from a synchronous social Q\&A system named IM-an-Expert (IMX), which routes incoming questions to candidate answerers via instant messaging \cite{2011-www_ryen_supporting_imx}. We obtained usage data for this system, from a deployment inside a large enterprise 
over a period of two years (May 2012 to April 2014). During that time, the system was used by about 5,000 subscribed users (the ``experts"), embedded in the network defined by the  organizational tree hierarchy of over $180,000$  nodes. These ``experts" each have a profile that had been inferred from their homepages, email distribution list subscriptions, and a self-described set of keywords describing their expertise. Users can pose a question in IMX via text, for example, \emph{Excel: How do I set default pivot table to ``Classic"?}. For each such question, the system assigns an expertise score to the subscribed users, in the range $(0,1]$, based on matching the content of question keywords against their profiles. While the deployed IMX system operates in a centralized manner with questions being routed to any expert, we simulated the system as only being able to locate experts using local search. 

\vspace{-2mm}
\subsection{Results}
\vspace{-1mm}

{\bf Utility acquired w.r.t set size.}
Figure~\ref{fig.result_syn_random_vary-cost_util},\ref{fig.result_syn_pa_vary-cost_util},\ref{fig.result_real_imx_vary-cost_util} illustrates the utility acquired as the specified upper bound on the selected set $\selectSet$ is increased, showing the robustness of \elgreedy. The difference in the rate at which the utility is acquired for the random graph (Figure~\ref{fig.result_syn_random_vary-cost_util}) and preferential attachment graph models (Figure~\ref{fig.result_syn_pa_vary-cost_util}) is in line with the theoretical results we had studied for worst-case as well as scale-free networks in  Theorem~\ref{theorem.general.l1} and Theorem~\ref{theorem.pa}, respectively. The results on the real-world IMX dataset in Figure\ref{fig.result_real_imx_vary-cost_util} shows a large positive gap in acquired utility for \elgreedy compared to the baselines. Note that the properties of the IMX dataset are very different from the two synthetic datasets, since the users are embedded in an organizational tree hierarchy.

{\bf Exposed network and varying feature distribution.}
The results in  Figure~\ref{fig.result_syn_random_vary-cost_visibility} shows that \elgreedy also maintains a good balance of the exposed network in comparison to  \baseval and to \basedeg (\emph{i.e.}, specifically optimizing for maximizing this exposure).  Figure~\ref{fig.result_syn_random_vary-probExpert} compares the different algorithms (with $|\selectSet|=50$), as the proportion of valued nodes in the network are increased in the first dataset.

{\bf Varying $\epsilon$ for \elgreedy.} 
Figure~\ref{fig.result_vary-epsilon}  shows the results for \elgreedy with varying $\epsilon$ on different three datasets, by restricting $|\selectSet|$ to $100$, $100$ and $200$ respectively. The results show the robustness of  \elgreedy compared to explore-only ($\epsilon=1$) and exploit-only ($\epsilon=0$) methods. The results plot the utility acquired divided by the size of selected set, (\emph{i.e.}, $\sfrac{\func(\selectSet)}{|\selectSet|}$), normalized to the same scale across datasets.

\vspace{-3mm}
\section{Conclusions and Future Work}\label{sec.conclusions}
\vspace{-1mm}
We formalized the process of selecting nodes in a network for information gathering, limited by local knowledge and visibility of the network. Our model captures many real-world applications that have been previously studied only under the assumption that the underlying network and complete set of nodes is known and accessible in advance. We developed a general algorithm \elgreedy that  provides a simple way to trade-off between network exploration and the exploitation of value in the exposed neighborhood.  We proved strong theoretical guarantees on its performance, and evaluated our approach on several simulated instances as well as using data collected from a real-world social Q\&A system. We see several interesting directions in which the presented work can be extended. We assumed that the utility of a node can be observed perfectly. Extending the proposed approach for the case of noisy/stochastic function evaluations would be practically useful. Designing algorithms that adjusts the rate of exploration $\epsilon$  and  take into account for some prior information available about the network structure and utility distribution is an interesting direction for future work. 

\vspace{2mm}
\noindent {\footnotesize \textbf{Acknowledgments.}
We'd like to thank Johan Ugander for helpful discussions. Andreas Krause was supported in part by a Microsoft Research Faculty Fellowship and ERC StG 307036.}
\vspace{-1.5mm}

\bibliographystyle{named}
\bibliography{ijcai15-information-gathering-netexp}
\clearpage
\onecolumn
\appendix 
{\allowdisplaybreaks
\section{Proof of Theorem 1}
\begin{proof}[\bf{Proof of Theorem~\ref{theorem.general.l2}}]
Let us denote the coin toss of the algorithm at each iteration  by a random variable $Y_i$ and its outcome as $y_i \in \{0,1\}$, with $y_i = 1$ corresponding to  the action $\actexplore$ and $y_i = 0$ corresponding to the action $\actexploit$. We shall separately analyze the sequence of $\actexplore$ actions and the sequence of $\actexplore$ actions. 
The selected set $\selectSet_i$ denotes the current set of nodes selected by the algorithm at the end of iteration $i$.

{\bf Analyzing the  $\actexplore$ actions:} Let us first analyze the  $\actexplore$ actions  of the algorithm. The $\actexplore$ actions of the algorithm aim towards building the connected dominating set of the graph. These actions simulate the greedy procedure of building connected dominating set with local $l_{deg}=2$ visibility in \cite{1998-_khuller_cds-algorithmica}, however with following two differences:
\begin{enumerate}
\item The $\actexplore$ actions are interleaved with the $\actexploit$ actions that can add some nodes to the selected set, and hence need to be taken into account during the analysis of building connected dominating set.
\item \elgreedy decides how many users to add in one round based on the following 
$$\Pi^* = \argmax_{\Pi^{\selectSet}_l: l \in [1 \ldots l_{deg}]} \frac{|\widetilde{\mathcal{N}}(\selectSet \cup \Pi^\selectSet_l,1) \setminus \widetilde{\mathcal{N}}(\selectSet,1)|}{l}$$

For the case of $l_{deg} = 2$, this essentialy means that the two nodes are added simultaneously, if their average exposure is higher than the exposure of the single node. This is in contrast and optimal compared to the \cite{1998-_khuller_cds-algorithmica} where two nodes are added as long as their joint exposure is higher than the single node,
\end{enumerate}

With these two differences, we analyze the execution of these $\actexplore$ actions, and prove below that a connected dominating set is built with size bounded by $\big(2 + 2\ln(\Delta_\graph) \big) $. Given that in expectation, only $\frac{1}{\epsilon}$ exploration steps are taken, there is additional factor of $\frac{1}{\epsilon}$ that appears.\\

Let $D^c_\graph \subseteq \groundSet$ denotes the minimal connected dominating set of the graph, of size $\gamma_\graph^c$. Let $z \in D^c_G$ denote one of the nodes in this set, and each such node $z$ covers a set of the nodes in $\groundSet$ denoted by the set $\groundSet^z$. If a node in $\groundSet$ is covered by multiple nodes in $D^c_\graph$, we randomly assign it to one of them. This way, $\{\groundSet^z\} \text{ } \forall \text{ } z \text{ in } D^c_\graph $ creates a partition of ground set $\groundSet$ into disjoint sets. Next, we associate colors with the nodes in $\groundSet$ as follows: \emph{i)} all the nodes are intitally white, \emph{ii)} whenever a node is added to the current set, it is turned black, \emph{iii)} all the newly exposed nodes of current set, \emph{i.e.,} immediate neighbors are turned gray.

We analyze the total size required to build connected dominating set using the charging analysis, similar to \cite{1998-_khuller_cds-algorithmica}. Each time, a node is added to the set (\emph{i.e.}, turned black), we charge by distributing this cost of increasing the set size to the newly exposed nodes (\emph{i.e.}, those turned gray). This charging is only done when  $y_i = 1$. We then bound the total charge per partition $\groundSet^z$ for $z \in D^c_\graph$. Note that, a node is charged only once as a node turns gray only once and hence the sum of these charges gives us the total number of the black nodes added (which is same as the size of the connected dominating set we would like to bound).

Let us now analyze the total charge that is placed on the nodes of $\groundSet^z$. We denote the number of white nodes in  $\groundSet^z$ at iteration $i$ by the notation $|\groundSet^z|_i$. Note that,  initially, all the nodes are white (\emph{i.e.,} $|\groundSet^z|_i$ = $|\groundSet^z|$) and $|\groundSet^z|_i$ eventually drops to zero once all the vertices in $\groundSet^z$ are turned gray or black. Consider the iteration $i^z_0$, when for the first time, any node in $\groundSet^z$ has changed color from white to gray color. Before  $i^z_0$, all the nodes in  $\groundSet^z$ remain white.  At this iteration $i^z_0$, up to two nodes in $\groundSet$ turned black, and $|\groundSet^z| - |\groundSet^z|_{i^z_0}$ number of nodes in $\groundSet^z$ have turned gray. The maximum charge that is applied on these nodes is given by: $\frac{2}{ |\groundSet^z| - |\groundSet^z|_{i^z_0}}$ (charge would be less, if only one node is turned black, \emph{i.e.,} numenator would be 1, or charge would be less, if additional nodes are turned gray in other partitions, \emph{i.e.,} demoninator would be higher).

After this initial $i^z_0$ iteration for this set, consider any iteration $j$. Now, the node $z$ is itself available to be selected (as part of the chain of size two), and it would expose atleast a set of size $|\groundSet^z|_{j-1}$. Hence the total exposure of the nodes (one or two) that are selected by the procedure in this iteration must be higher than this. This implies that if any node is turned gray from $\groundSet^z$ in this iteration, the charge is bounded by  $\frac{2}{|\groundSet^z|_{j-1}}$.

The total charge that is placed on the nodes in set $\groundSet^z$ can be bounded as follows, taking into account that a charge is placed only for the explore actions (\emph{i.e.,} $y = 1$):

\begin{align*}
&{y_{i^z_0}} \cdot \frac{2 \cdot (|\groundSet^z| - |\groundSet^z|_{i^z_0})}{(|\groundSet^z| - |\groundSet^z|_{i^z_0})}  + {y_j} \sum_j \frac{2 \cdot (|\groundSet^z|_{j-1} - |\groundSet^z|_{j})}{|\groundSet^z|_{j-1}} \\
&\leq 2 + {y_j} \cdot 2 \cdot \sum_j \frac{|\groundSet^z|_{j-1} - |\groundSet^z|_{j}}{|\groundSet^z|_{j-1}} \\
&\leq 2 + 2 \cdot \sum_j \frac{|\groundSet^z|_{j-1} - |\groundSet^z|_{j}}{|\groundSet^z|_{j-1}} \\
&\leq 2 + 2\ln(\Delta_\graph)
\end{align*}

In second and third steps above, the binary variable $y$ is replaced by the value of $1$. Now, given that there are total of $\gamma^c_G$ such partitions, hence the total charge that can be placed (equivalent the total number of black nodes selected) is given by $\big(2 + 2\ln(\Delta_\graph) \big) \cdot  \gamma^c_\graph$. 

\vspace{2mm}

{\bf Analyzing the  $\actexploit$ actions:} Let us consider the phase when the connected dominating set is already built (\emph{i.e.}, there are no more remaining white nodes). In the worst case setting, if the selected set doesn't provide the required utility yet, from this phase onwards, all the nodes are directly accessible to the algorithm as it would be for a centralized algorithm with full visibility and without any connectivity constraints.  The $\actexploit$ actions of the algorithm aim towards maximizing the set function as in a standard greedy approach of submodular function maximization (\cite{2012-survey_krause_submodular}), however with following two differences:
\begin{enumerate}
\item The $\actexploit$ actions are interleaved with the $\actexplore$ actions that can add some nodes to the selected set. However, the greedy algorithm for submodular function maximization is oblivious to this addition of other nodes and can be ignored from the analysis.
\item The algorithm has already selected a set of nodes by the time this phase starts (when connected dominating set is built). However, again, the analysis of the greedy algorithm for submodular function maximization is oblivious to the starting set that may have already been selected by the procedure.
\end{enumerate}

With these two differences, the execution of these $\actexploit$ actions, starting from the phase when connected dominating set is built, leads to the bound of $ |\opt| \cdot \ln(\frac{1}{\beta})$ on the number of additional nodes that could be added to the set until the required utility value is achieved. Given that in expectation, only $\frac{1}{1 - \epsilon}$ exploitation steps are taken, there is additional factor of $\frac{1}{1 - \epsilon}$ that appears.  In fact, a tighter analysis can be done because of the following optimization trick in the algorithm that, when the full network is exposed (or dominating set for the network is already built), we set $\epsilon=0$ (see Step~\ref{alg.elgreedy.eps0} of Algorithm~\ref{alg.elgreedy}).
\end{proof}

\section{Proof of Theorem 2}

\begin{proof}[\bf{Proof of Theorem~\ref{theorem.general.l1}}]
The proof of this theorem follows exactly along the same arguments as that of Theorem~\ref{theorem.general.l2}. However, for $l_{deg}=1$, the $\actexplore$ action has a randmozization step involved as discussed in Section~\ref{sec.mechanism}. To recall, as shown in the seminal work of \cite{1998-_khuller_cds-algorithmica}, a simple deterministic procedure for $l_{deg}=1$ that add users greedily based on maximizing the ``exposure" may have worst case cost of $|\groundSet|$. Recently, this barrier of $l_{deg} = 1$ has been resolved by \cite{2012_power-of-local-info} using a simple randomization technique. Specifically, after a user $v$ is added, another randomly selected user from her neighborhood $\mathcal{N}(v,1)$ that is newly exposed is also added. This is illustrated in the Step~\ref{alg.elgreedy.random} of Algorithm~\ref{alg.elgreedy}.

In fact, the analysis of this new randomizied technique proposed in \cite{2012_power-of-local-info} is inspired from the \cite{1998-_khuller_cds-algorithmica} and follows similar techniques for the analysis. Our results for this theorem follows along the same analysis as done in proof of Theorem~\ref{theorem.general.l2}. We separately analyze the $\actexplore$ and $\actexploit$ actions, ensure that interleaving these two actions doesn't affect the analysis, and then combine these two actions by multiplying the two terms in the bounds by $\frac{1}{\epsilon}$ and $\frac{1}{1 - \epsilon}$ respectively.
\end{proof}

\section{Proof of Theorem 3}

\begin{proof}[\bf{Proof of Theorem~\ref{theorem.general.lower}}]
The theorem states the lower bound for the general settings considered in Section~\ref{sec.analysis.general} in the worst case. For any bounded $l_{deg}$ and $l_{val}$, there exists a problem instance for which any feasible policy will need to select a set of size at least  $|\selectSet| \geq \max{} \Big(\gamma^c_\graph,   \text{ }   |\opt|\Big)$.

The proof of Theorem~\ref{theorem.general.lower} follows from the arguments of Section~\ref{sec.mechanism.methodology} and by crafting a worst-case instance of graph structure and distribution of node values for any given policy. It is clear that any algorithm will require a set of  size at least  $|\opt|$, by the definition of $\opt$ from Section~\ref{sec.model}. Recall that $\opt$ is the algorithm operating without any computational constraints, with global visbility and without any contraints on connectivity of the set. We shall now show that the dependency on $\gamma^c_\graph$ is unavoidable by creating a worst case problem instance for any given algorithm as follows. 

Consider a ground set $\groundSet$ of nodes. All the tasks originate from a fixed node $v_o \in \groundSet$. All the nodes $\groundSet$ posses $0$ utility, except a special valued-node $v_x$ possessing utility of $1$. Hence, the goal of any algorithm is to locate $v_x$ starting from $v_o$. Consider any deterministic algorithm which adds the nodes to the set in a particular order. We split the execution of this algorithm in two phases. Phase 1 consists of the set of nodes which are added to the set, until a connected dominating set is formed and let us denote this set of nodes added to the set as $\selectSet_1$. Phase 2 consists of the set of nodes which are added after a connected dominating set is formed, and we denote this set of nodes as $\selectSet_2$. The final set output by the algorithm is given by $\selectSet= \selectSet_1 \cup \selectSet_2$. For this algorithm, we pick the valued node $v_x \in \groundSet \setminus  \selectSet_1$. In order words, for any algorithm, a worst case distribution of the node values can be set up, such that the single valued node $v_x$ cannot be located without constructing the connected dominating set. This gives us the worst case lower bound of $\gamma^c_\graph$ for any determinisitc algorithm. Similarly, for any randomized algorithm or policy, it can be shown to have a dependency on $\gamma^c_\graph$, with probability bounded away from zero.
\end{proof}

\section{Proof of Theorem 4}

We begin by providing more details of the model introduced in Section~\ref{sec.analysis.pa} and then proving  Theorem~\ref{theorem.pa}.
\subsection{Details of the model for realistic settings}\label{sec.analysis.pa.details}

{\bf Skill distribution.}
The nodes are associated with subsets of a features from the set $\featureSet$ of size $|\featureSet|$. Considering the specific domain of co-authorship networks, a feature $x \in \featureSet$ could, e.g., be an indicator variable denoting whether a user has published a paper in an AI conference. For each feature $x \in \mathcal{X}$, let the set of nodes possessing that feature be given by  $\groundSet_x$ and corresponding probability by $\theta_x = \frac{|\groundSet_x|}{|\groundSet|}$. We assume that the features are independently distributed.  Under independence assumption, we can state that if we take any node $v$ and consider its $m$ outgoing links, then the probability that at least one of nodes from these $m$ links poses feature $x$ is at least $(1 - (1- \theta_x)^M)$.

{\bf Characteristic properties of the graphs.}
We consider each feature $x \in \mathcal{X}$ as a ``social dimension" that induces a network given by $\graph_x=(\groundSet_x,\edges_x)$. In this model we assume that each network $\graph_x$ is formed by preferential attachment process \cite{1999-science_emergence-scaling}, where nodes $v \in \mathcal{\groundSet}_x$ arrive over time, and each node on arrival forms $m_x$ links to existing nodes with probability proportional to the degree of those nodes. This network leads to power law distribution, often seen in the collaborative networks such as co-authorship graphs or online social network. The final network that we observe is obtained by an overlay of these networks, given by $\graph = \cup_{x \in \mathcal{X}} \graph_x$. For a node $v$, we denote the value of its feature $x \in \featureSet$ as $x_v$ and  is proportional to the rank order dictated by the degree of node $v$ in graph $\graph_x$, scaled to have bounded support of $(0,1]$. This notion of feature values essentially captures the ``authority" of a node over a particular feature (for instance, a node gains expertise in ``AI" if it has high degree of connections for this particular network). Note that, we consider undirected networks, and hence incoming or outgoing degree is same.

{\bf Characteristics properties of utility function.}
We now characterize the utility function properties. For a given task $\task$, we consider a separable utility function given by: $\func(\selectSet) = \sum_{x \in \featureSet} w^x \cdot \func^x(\selectSet)$,  where $w^x$ denotes the weight of function $\func^x$. The weights are normalized, \emph{i.e.}, $\sum_{x \in \featureSet} w^x = 1$. Here, the function $\func^x$ depends only on nodes' features  $x$, and its value is bounded between $[0,1]$. There are set of ``irrelevant" features for which task $\task$ carries no utility, \emph{i.e} $w_\task^x$ = 0. For other ``useful" features for which $w_\task^x > 0$, function $\func_\task^x$ carries positive, non-zero utility for every node possessing non-zero value of that feature. The maximum value of function $\func^x(\selectSet) = 1$ is achieved after including a top valued node with feature $x$.  This is motivated by real-world settings where the goal is essentially to find few experts for each of the required skills for a task (such as a collaborative project in academia).

{\bf Characteristics properties of \opt.}
For a given task, consider the function associated with any specific useful feature $x$ for which $w_\task^x > 0$. As discussed above, then the maximum value of function $\func_\task^x(\selectSet) = 1$ will be achieved after including a top valued node with feature $x$.  Hence, $\opt$ contains one top valued node for each of the features for which $w_\task^x > 0$.


\subsection{Proof}
\begin{proof}[\bf{Proof of Theorem~\ref{theorem.pa}}]
Theorem~\ref{theorem.pa} is proved using the results on the navigation properties of ``small-world" networks \cite{2003-_mathematical-results-pa,2012_power-of-local-info}. Compared to the general settings of Theorem~\ref{theorem.general.l2} and~\ref{theorem.general.l1}, the ``small-world" networks are much easier to navigate as can be seen in the polylogarithmic bounds.

Consider a task $T$ originating from node $v_T^o$ possessing feature $x^o$. Let the goal is to find at least one top valued node for each of the features with  $w_\task^x > 0$ . 
Theorem~\ref{theorem.pa} states the main results for these settings, that is, with probability of at least $1 - o(1)$, \elgreedy (with $l_{deg}=1$, $l_{val}=1$) terminates  with set $\selectSet$ achieving  utility of at least $(1 - \beta) \cdot \quota$, with the following bound on the size of $\selectSet$ in expectation,

\begin{align*}
\mathbb{E}[|\selectSet|] \leq \Big(\frac{1}{\epsilon} \!\cdot\! O\big(\ln^4(|\groundSet_{x^o}|)\big)\Big)\! +\! \Big(\frac{1}{1 - \epsilon} \!\cdot\! O\big(\sum_{x \in \featureSet} \ln^4(|\groundSet_x|) + |\opt| \big)\Big)  
\end{align*}

\noindent
The main ideas behind the proof of the above theorem are illustrated in Figure~\ref{fig.theorem4} and discussed below. For the ease of putting forth the ideas, let us first consider only one useful feature/skill.

\begin{figure*}[h]
\centering
\includegraphics[width=0.6\textwidth]{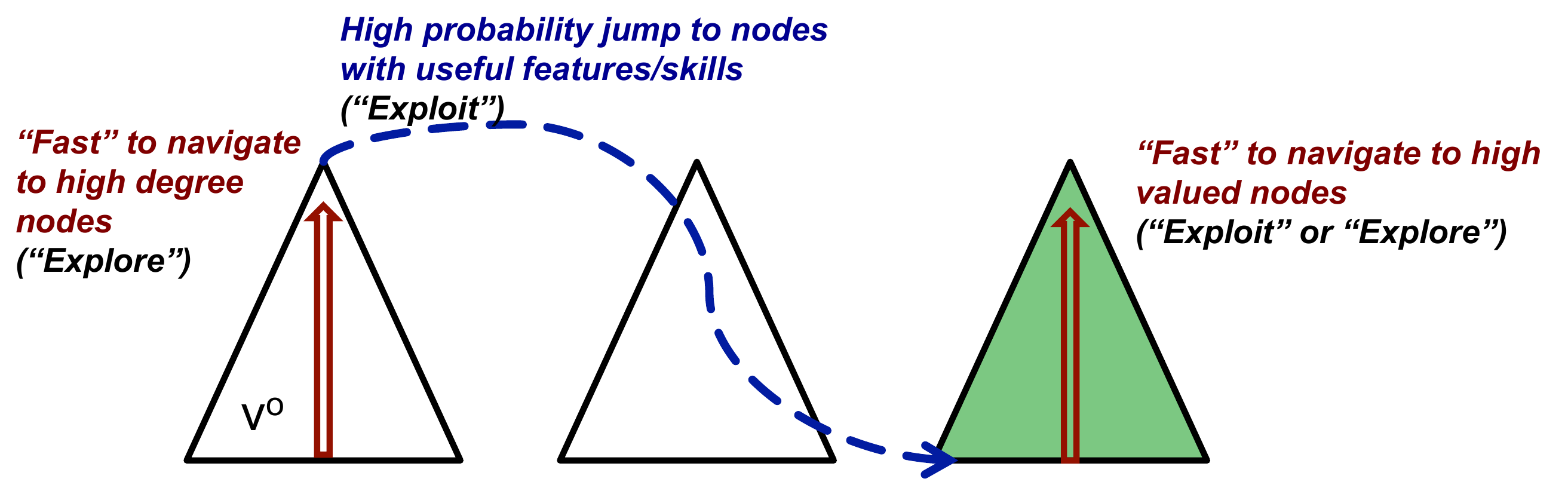}
\vspace{-2mm}
\caption{Illustration of the keys ideas of navigating to high valued nodes in the specific settings of preferential attachment graphs. Each triangle here represents a graph associated with a feature, with higher degree nodes of that feature located towards the top of the triangle. In this example, there are total of three features, and one useful feature shown by the green triangle. }
\label{fig.theorem4}
\end{figure*}

{\bf Reaching high degree nodes in the graph $\graph_{x^o}$.} The first term in the summation $\Big(\frac{1}{\epsilon} \!\cdot\! O\big(\ln^4(|\groundSet_{x^o}|)\big)\Big)$  is attributed to the process of identifying high degree nodes among the nodes possessing feature $x^o$. Let us denote one such high degree node as $r_{x^o}$. The exploration step of the algorithm will navigate to the node $r_{x^o}$ within $O\big(\ln^4(|\groundSet_{x^o}|)\big)$ steps such that the $\Delta_{\graph_{x^o}}(r_{x^o}) \geq \frac{m_{x^o} \cdot  \sqrt{|\groundSet_{x^o}|}}{\ln(|\groundSet_{x^o}|}$ (where $\Delta_G(v)$ denotes the degree of vertex $v$ in the graph \graph). This follows from the navigation properties of the preferential attachment model of networks \cite{2012_power-of-local-info}. Consider the process of navigating the graph $G_{x}$ for a feature $x$, starting from any node with feature $x$. If we follow the maximum degree node in the $l_{deg}=1$ local visibility, we reach a node $r_x$ within $O\big(\ln^4(|\groundSet_x|)\big)$ steps such that $\Delta_{\graph_x}(r_x) \geq \frac{1}{\ln^2(|\groundSet_x|)} \cdot \Delta_{\graph_x}$, this holds with probability at least $1 - o(1)$ (where the $o$ notation is w.r.t to size of the graph). Furthermore, with probability at least $1 - o(1)$, the maximum degree of any node in $\graph_x$, denoted by $\Delta_{\graph_x}$ is bounded by $\Delta_{\graph_x} \leq m_x \cdot  \sqrt{|\groundSet_x|} \cdot \ln(|\groundSet_x|)$ (\cite{2003-_mathematical-results-pa}). 

Given that in expectation, only $\frac{1}{\epsilon}$ exploration steps are taken, the first term represents the bound on the number of nodes that could be added to the selected set  towards reaching $r_{x^o}$. 

{\bf Jumping to some nodes with useful features.}  Then, the exploitation step would ensure that we can make a jump from $r_{x^o}$ to the nodes with useful features  (\emph{i.e.}, $w_\task^x > 0$), with high probability.  Consider a useful feature $x$, given the independence assumption of the feature distribution, the algorithm can jump to nodes with feature $x$, with probability at least: $1 - (1- \theta_x)^{\Delta_{\graph_{x^o}}(r_{x^o})}$, and this is $1 - o(1)$.


{\bf Reaching high degree nodes with useful features.} Both the exploration and exploitation step would help with this process. However, the exploration step may still continue adding higher degree nodes in the graph $G_{x^o}$ or other features that do not provide any further utility, hence the second term in bound of Theorem~\ref{theorem.pa} has a factor of $\frac{1}{1 - \epsilon}$.
Based on the results of  \cite{2012_power-of-local-info}, consider starting from any node and navigating graph $G_x$, if we follow the maximum degree node in the $l_{deg}=1$ local visibility, we reach a  small constant number of high degree $(1 + \kappa)$ nodes, denoted by $R_x^\kappa$, within $O\big(\ln^4(|\groundSet_x|) + \kappa \big)$ steps such that $\Delta_{\graph_x}(r_x) \geq \frac{1}{\ln^3(|\groundSet_x|)} \cdot \Delta_{G_x}$ for $r_x \in R_x^\kappa$, this holds with probability at least $1 - o(1)$. As per the power law distribution of the degrees, this set $R_x^\kappa$ of $(1 + \kappa)$ highly-skilled nodes are within top $1- \beta$ percentile top rank for the feature $x$, with probability $1 - o(1)$.

When considering only one useful feature, putting these ideas together provide us the main results of Theorem~\ref{theorem.pa} for these settings, which holds with probability of at least $1 - o(1)$. The same ideas also carry over when there are multiple useful features. This follows from the fact that once  $1+\kappa$ high valued nodes of one particular feature have already been added to the set, then the nodes of this feature will not provide any further marginal gain, and hence the exploitation step will jump to nodes of another useful feature that is required by the task. The second term in the statement of theorem above indeed needs to be summed only over the useful features.  This summation over all the features $\featureSet$ in the second term represents the worst case setting when all the features are useful and required by the task.

 In our specific problem setup in Section~\ref{sec.analysis.pa.details} and the results Theorem~\ref{theorem.pa}, the utility function requires only one top valued node per useful feature, hence $\kappa$ can be set to $0$ in the above bounds.
\end{proof}

\section{Proof of Theorem 5}

\begin{proof}[\bf{Proof of Theorem~\ref{theorem.pa.lower}}]

Next, we state the lower bound in Theorem~\ref{theorem.pa.lower}, which follows from the expected diameter of the graph obtained from preferential attachment \cite{2004-_scale-free-preferential}.

The theorem states the lower bound for the specific model considered in Section~\ref{sec.analysis.pa}, with further details provided in  Section~\ref{sec.analysis.pa.details}. The results follow from the expected diameter of the graph obtained from preferential attachment \cite{2004-_scale-free-preferential}. We can create the problem instance as follows. Let there be only one unique feature $x$ (\emph{i.e.}, $|\featureSet| = 1$ and let all the nodes $\groundSet$ possess this feature,  \emph{i.e.}, $\groundSet = \groundSet_x$ with all the nodes possessing this feature. All the tasks originate from  nodes for which the degree is one. 

The expected diameter of the graph obtained from the preferential attachment process is given by $O\Big(\frac{\ln(|\groundSet|)}{\ln\ln(|\groundSet|)}\Big)$ \cite{2004-_scale-free-preferential}. Any algorithm or policy, (even if it has full knowledge of the network, \emph{i.e.}, with unbounded $l_{deg}$, $l_{val}$) will have the expected size of selected set at least that of the diameter as it represents the minimal path length to connect the nodes from where task originates (those with degree of one in this problem instance) to the highest valued node.
\end{proof}
}

\end{document}